\documentclass[11pt]{article}

\usepackage[dvipdfmx]{graphicx}

\usepackage{url}
\usepackage{authblk}

\usepackage{amsmath} 
\usepackage{txfonts}
\usepackage{color}
\usepackage{url}
\usepackage{here}
\usepackage{comment}
\usepackage{bm}
\usepackage{algorithm}
\usepackage{algorithmic}
\usepackage{lineno}
\usepackage{ulem}
\usepackage[margin=25mm]{geometry}

\newcommand{\argmin}{\mathop{\rm argmin}\limits}

\title{Extended dynamic mode decomposition with dictionary learning \\using neural ordinary differential equations} 

\author[1]{Hiroaki Terao\thanks{hiroaki.terao@ist.osaka-u.ac.jp}}
\author[1]{Sho Shirasaka}
\author[1]{Hideyuki Suzuki}

\affil[1]{Graduate School of Information Science and Technology, 
Osaka University\\1-5 Yamadaoka, Suita, Osaka 565-0871, Japan}

\begin{document}

\maketitle

\begin{abstract} 
Nonlinear phenomena can be analyzed via linear techniques using operator-theoretic approaches. Data-driven method called the extended dynamic mode decomposition (EDMD) and its variants, which approximate the Koopman operator associated with the nonlinear phenomena, have been rapidly developing by incorporating machine learning methods. Neural ordinary differential equations (NODEs), which are a neural network equipped with a continuum of layers, and have high parameter and memory efficiencies, have been proposed. In this paper, we propose an algorithm to perform EDMD using NODEs. NODEs are used to find a parameter-efficient dictionary which provides a good finite-dimensional approximation of the Koopman operator. We show the superiority of the parameter efficiency of the proposed method through numerical experiments.

\end{abstract}



\section{Introduction} 
In recent years, interest in data analysis has increased with development of computer hardware and accelerated accumulation of various data such as sensor data and statistical data. Various problems and phenomena of importance are represented by data of time series such as weather history, stock price fluctuations, and web traffic. Thus time series analysis is an important problem in the fields of science and engineering~\cite{Ikeguchi,Kantz}.
\par
In time series analysis, there are two types of target phenomena. One is linear and the other is nonlinear. In this study, we analyze nonlinear phenomena, because they are ubiquitous in the real world. Nonlinear dynamical systems exhibit complex behavior such as multi-stabilities, limit-cycle oscillations and chaos, 
which have been utilized to explore innovative frontiers in information processing systems, such as distributed intelligent systems, chaotic communication systems, hardware accelerators and neuromorphic computers ~\cite{Bullo,Tse,Inagaki,Maass}. In general, nonlinear systems do not admit simple straightforward analysis as in the cases of linear ones.
\par
The Koopman operator is a linear operator which describes observations along trajectories of dynamical systems~\cite{Budisic,Susuki}. 
Though in general it is infinite-dimensional, it fully captures the underlying nonlinear dynamical systems. 
Nonlinear dynamical systems can be analyzed using established linear techniques through the operator. 

A crucial step in practical applications of the Koopman operator theoretic framework is a finite-dimensional approximation of the operator via estimating its eigenpairs. 
This can be performed in a data-driven way using a method called the dynamic mode decomposition (DMD)~\cite{Kutz}. 
Williams {\it et al.}~proposed the Extended DMD (EDMD) which uses an expanded set of observables introduced a priori, called dictionary, in order to make a more accurate approximation of the Koopman operator~\cite{Williams}.  
Machine learning techniques of deep neural networks (DNNs)~\cite{Li,Yeung,Lusch,Takeishi,Mardt} have been introduced so as to improve upon the EDMD algorithm. 
%
In EDMD with dictionary learning (EDMD-DL)~\cite{Li}, multilayer perceptrons (MLPs) are used as a trainable set of the dictionary, which enables us to find an efficient representation of the approximated Koopman operator. 

Recently, neural ordinary differential equations (NODEs) have been proposed as continuous-depth models of MLPs~\cite{Chen}. 
NODEs can be trained with a constant memory cost which does not depend on the number of layers and with a smaller set of parameters than that of the conventional MLPs. 

The purpose of this study is to propose a computationally efficient algorithm for learning a finite-dimensional approximation of the Koopman operator using NODEs. 
While sharing the same dictionary learning strategy as that of the original EDMD-DL~\cite{Li}, the MLPs in the algorithm are replaced by the NODEs to improve the computational performance in this work. 
Using two nonlinear systems, the Duffing equation and the Kuramoto-Sivashinsky equation, as examples, we demonstrate that the Koopman operator approximated by our proposed method provides a good prediction of the complex nonlinear behavior and that the NODE-based EDMD-DL outperforms the conventional MLP-based one in terms of parameter efficiency.

\section{Methods}
In this section, we first introduce the Koopman operator, then give a brief overview of the EDMD-DL and the NODEs, and finally propose the NODE-based EDMD-DL. 

\subsection{Koopman operator\label{sec. koopman}} 
Nonlinear phenomena are found everywhere, but, in general, it is difficult to analyze them in a uniform way. 
However, we can analyze such nonlinear dynamical systems via linear techniques using the Koopman operator. 
Here, the term "linear techniques" means an exact linearization, not an approximation. 
In this subsection, we explain how we can analyze nonlinear dynamical systems via linear techniques using the Koopman operator.

Consider a nonlinear dynamical system $\mathbf{x}_{n + 1}=\mathbf{F}(\mathbf{x}_{n})$, where $\mathbf{x}_{n} \in \mathcal{M}$ is a state variable, $\mathcal{M}$ is a state space, and $\mathbf{F}: \mathcal{M} \to \mathcal{M}$ is a map which defines a law of time-evolution. 
The Koopman operator $\mathcal{K}$ is defined as 
\begin{align}
\mathcal{K}\mathbf{g}(\mathbf{x}_{n}) := (\mathbf{g} \circ \mathbf{F})(\mathbf{x}_{n}), \label{eq. defkop}
\end{align}
where $\circ$ denotes a function composition and $\mathbf{g}:\mathcal{M} \to \mathbb{C}^d$, called an observable, is a $d$-dimensional vector valued function defined on the state space of the system. 
Eq.~\eqref{eq. defkop} implies
\begin{align}
\mathcal{K}\mathbf{g}(\mathbf{x}_{n}) = \mathbf{g}(\mathbf{F}(\mathbf{x}_{n})) = \mathbf{g}(\mathbf{x}_{n + 1}), \label{eq. defkop2}
\end{align}
Hence, the Koopman operator $\mathcal{K}$ describes sequential observations along trajectories of dynamical systems. 
The Koopman operator $\mathcal{K}$ is a linear operator but it typically requires infinite-dimensional analysis because it acts on a space of functions. It can be shown that $\mathcal{K} $ is linear as follows. Let $\mathbf{g_1}:\mathcal{M} \to \mathbb{C}^d$ and $\mathbf{g_2}:\mathcal{M} \to \mathbb{C}^d$ be observables, respectively. Let $\alpha_1 \in \mathbb{C}$ and $\alpha_2 \in \mathbb{C}$ be constants, respectively. From Eq.~\eqref{eq. defkop}, the following equation holds:
\begin{align}
\mathcal{K}(\alpha_1\mathbf{g_1}(\mathbf{x}_{n}) + \alpha_2\mathbf{g_2}(\mathbf{x}_{n})) = [(\alpha_1\mathbf{g_1} + \alpha_2\mathbf{g_2})\circ \mathbf{F}] (\mathbf{x}_{n}) = \alpha_1\mathcal{K}\mathbf{g_1}(\mathbf{x}_{n}) + \alpha_2\mathcal{K}\mathbf{g_2}(\mathbf{x}_{n}).
\end{align}
Since $\mathcal{K}$ is linear, we consider the eigenpairs of the operator:
\begin{align}
\mathcal{K} \phi_{k} = \mu_{k}\phi_{k}, \label{eq. eigen}
\end{align}
where $\mu_k \in \mathbb{C}$ and $\phi_k \in \mathcal{M}\to \mathbb{C}$ are called the Koopman eigenvalue and the Koopman eigenfunction, respectively. 
\par
Assume that $\mathbf{g}$ can be expressed as a linear combination of eigenfunctions $\{\phi_{1}, \phi_{2}, ...\}$ as follows: 
\begin{align}
  \mathbf{g} = \sum_{k = 1}^{\infty} \mathbf{m}_{k}\phi_{k}, \label{eq. series}
\end{align}
where $\mathbf{m}_{k} \in \mathbb{C}^d$, called the Koopman mode, is a coefficient vector of the associated $k$-th Koopman eigenfunction. 
From Eqs.~(\ref{eq. defkop}, \ref{eq. eigen}), we obtain 
\begin{align}
\phi_{k}(\mathbf{x}_{n}) = \mu_{k}\phi_{k}(\mathbf{x}_{n - 1}).
\end{align}
These relations lead to 
\begin{align}
  \mathbf{g}(\mathbf{x}_{n}) &= \mathcal{K}\mathbf{g}(\mathbf{x}_{n - 1})
   = \sum_{k = 1}^{\infty} \mathbf{m}_{k}(\mathcal{K}\phi_{k})(\mathbf{x}_{n - 1}) \notag \\ 
   &= \sum_{k = 1}^{\infty} \mu_{k}\mathbf{m}_{k}\phi_{k}(\mathbf{x}_{n - 1}).  
\end{align}
A recursive argument provides 
\begin{align}
  \mathbf{g}(\mathbf{x}_{n}) = \mathcal{K}^n\mathbf{g}(\mathbf{x}_{0})  =  \sum_{k = 1}^{\infty} \mu_{k}^{n}\mathbf{m}_{k}\phi_{k}(\mathbf{x}_{0}). \label{eq. series2}
\end{align}
Here, letting $\mathcal{K}^n$ act on the observable is equivalent to an operation that steps forward in time by $n$ steps. 
Thus, if the Koopman modes and eigenpairs are obtained, we can make predictions of time evolution of the observable. 
Practically, only a finite number of the eigenpairs and the modes are estimated computationally. Hence the predictions are made using a truncated series of Eq.~(\ref{eq. series2}) as follows:  
\begin{align}
  \mathbf{g}(\mathbf{x}_{n}) \approx \sum_{k = 1}^{M} \mu_{k}^{n}\mathbf{m}_{k}\phi_{k}(\mathbf{x}_{0}),  \label{predictive_Koopman}
\end{align}
where $M$ is the number of eigenpairs and modes. 
\par
In this paper, time series prediction capacity of the Koopman operator framework is mainly discussed. However, the eigendecompostion of the Koopman operator is not only useful for the prediction tasks, but also provides us transparent interpretation of the underlying dynamical processes. For example,  for a broad classes of non-chaotic systems, $\mu_k$s are generated by linear combinations with natural number coefficients of the characteristic exponents~\cite{Mezic}. Hence, the Koopman eigenvalues can be an indicator of (possibly catastrophic) critical phenomena~\cite{Scheffer}. Also, the Koopman eigenfunctions associated with the characteristic exponents located in the close vicinity of the unit circle describe transient dynamical behavior that is relatively persistent over time. Therefore, such eigenpairs lead to a systematic dimension reduction of the dynamical systems~\cite{Shirasaka,Monga}. Moreover, as illustrated in Subsec.~\ref{Duffing} in this paper, the Koopman eigenfunctions provide information on a global phase space structure such as basins of attraction. Once the eigendecomposition is obtained, such nontrivial dynamical properties are made already apparent even without simulating the prediction model (\ref{predictive_Koopman}), in contrast to the black-box prediction models, which can be powerful but deny clear interpretation.

\subsection{Extended dynamic mode decomposition} 
EDMD gives a finite-dimensional approximation of the Koopman operator using a priori introduced dictionary of observables. 
This subsection describes the EDMD algorithm and its relation to the Koopman operator.

First of all, a time series data $\mathbf{X} \in \mathbb{R}^{d\times (N+1)}$ is prepared: 
\begin{align}
\mathbf{X} = [\mathbf{x}_{0}, \mathbf{x}_{1}, ..., \mathbf{x}_{N}]. \label{eq. dataset}
\end{align}
Next, a dictionary of observables $\mathbf{\Psi}(\mathbf{x}) \in \mathbb{R}^{1\times M}$ is introduced as 
\begin{align}
  \mathbf{\Psi}(\mathbf{x}) = [\psi_{1}(\mathbf{x}), \psi_{2}(\mathbf{x}), ..., \psi_{M}(\mathbf{x})]. 
\end{align}
%
Let $\mathbf{B} \in \mathbb{R}^{M\times d}$ be an weight matrix, and assume that the observable $\mathbf{g}$ can be expressed as a linear combination of elements of a dictionary $\{\psi_{1}, \psi_{2}, ...\}$ as follows: 
\begin{align}
\mathbf{g} = \mathbf{B}^{\mathrm{T}}\mathbf{\Psi}^{\mathrm{T}}. \label{B}
\end{align}
Using the data set and the dictionary, we calculate a matrix $\mathbf{K} \in \mathbb{R}^{M\times M}$ defined as 
\begin{align}
\mathbf{K} := \mathbf{G}^{+} \mathbf{A}, \label{K}
\end{align}
where
\begin{align}
\mathbf{G} &= \frac{1}{N}\sum_{n = 0}^{N - 1} \mathbf{\Psi}(\mathbf{x}_{n})^{\mathrm{T}}\mathbf{\Psi}(\mathbf{x}_{n}), \label{G} \\
\mathbf{A} &= \frac{1}{N}\sum_{n = 0}^{N - 1} \mathbf{\Psi}(\mathbf{x}_{n})^{\mathrm{T}}\mathbf{\Psi}(\mathbf{x}_{n + 1}). \label{A}
\end{align}
Here, $+$ is the pseudo-inverse and $\mathrm{T}$ denotes the transposition. 
$\mathbf{K}$ is a finite-dimensional approximation of the Koopman operator $\mathcal{K}$~\cite{Williams}. 
A subset of the Koopman eigenvalues can be approximated using that of $\mathbf{K}$. 
Let its $k$-th left eigenvector be ${\bm \xi}_k$ and the right one be ${\bm \zeta}_k$. 
The $k$-th left eigenvector is scaled so that ${\bm \xi}_{k}^{\ast} {\bm \zeta}_{k} = 1$, where $\ast$ denotes the Hermitian transpose.
The Koopman modes are obtained using the following equation:  
\begin{align}
\mathbf{m}_{k} = ({\bm \xi}_{k}^{\ast}\mathbf{B})^{\mathrm{T}}, \label{m}
\end{align}
Also, the EDMD approximation of the Koopman eigenfunction is
\begin{align}
\phi_{k}(\mathbf{x}) = \mathbf{\Psi}(\mathbf{x}){\bm \zeta}_{k}. \label{phi}
\end{align}
Thus, from Eqs.~(\ref{predictive_Koopman},\ref{m},\ref{phi}), $\mathbf{g}(\mathbf{x}_{n})$ is approximated as follows:
\begin{align}
  \mathbf{g}(\mathbf{x}_{n}) \approx \sum_{k = 1}^{M} \mu_{k}^{n}({\bm \xi}_{k}^{\ast}\mathbf{B})^{\mathrm{T}}\mathbf{\Psi}(\mathbf{x}_0){\bm \zeta}_{k}. \label{predictive_EDMDo}
\end{align}
In this study, we set the observable $\mathbf{g}$ as the identity map, i.e., $\mathbf{x}_{n} = \mathbf{g}(\mathbf{x}_{n})$. 
In this case, the evolution of the state variable $\mathbf{x}_{n}$ can be predicted by the following equation:
\begin{align}
  \mathbf{x}_{n} \approx \sum_{k = 1}^{M} \mu_{k}^{n}({\bm \xi}_{k}^{\ast}\mathbf{B})^{\mathrm{T}}\mathbf{\Psi}(\mathbf{x}_0){\bm \zeta}_{k}. \label{predictive_EDMD}
\end{align}

The classical DMD~\cite{Tu} corresponds to an EDMD with a dictionary composed of the standard projection maps ${\bm \pi}_k(\mathbf{x}) := {\bm e}_k^{\mathrm T}\mathbf{x}$, where ${\bm e}_k$ is the $k$-th unit vector~\cite{Williams}. 
The dictionary rarely captures the nonlinear nature of the underlying dynamical system and it can only span at most $d$-dimensional invariant subspace of the Koopman operator. 
In the EDMD, we can introduce suitable basis functions as a dictionary, such as radial bases for periodic systems. Moreover, in principle, we can decompose an arbitrarily large Koopman invariant subspace by introducing complete bases. Therefore, the EDMD provides a richer and more accurate understanding of the Koopman operator.

\subsection{EDMD-dictionary learning} 
In EDMD, good approximation of the Koopman operator might not be achieved unless the dictionary is properly selected since only finite number of basis functions are available. 
EDMD-DL provides a good dictionary in a systematic manner using DNNs such as MLPs.
%
In EDMD-DL, DNNs are used as a trainable dictionary which adaptively express a space spanned by a set of Koopman eigenfunctions, called the Koopman invariant subspace. 
The learning strategy described in this section is originally proposed in~\cite{Li}. We also employ this kind of dictionary learning cycle in this study. 
\par
The DNNs have parameters called weights and biases, which we denote as ${\bm \theta} \in \Theta$. 
The trainable dictionary $\mathbf{\Psi}$ is parametrized by ${\bm \theta}$, i.e., $\mathbf{\Psi}(\mathbf{x}) = \mathbf{\Psi}(\mathbf{x};{\bm \theta})$. 
The dictionary is trained so as to minimize the following loss function $J$:   
\begin{align}
&(\mathbf{K}, {\bm \theta}) = \argmin_{(\widetilde{\mathbf{K}}, \widetilde{\bm \theta})}{J(\widetilde{\mathbf{K}}, \widetilde{\bm \theta})} \notag \\ 
&= \argmin_{(\widetilde{\mathbf{K}}, \widetilde{\bm \theta})}{\sum_{n = 0}^{N - 1} \left\|\mathbf{\Psi}(\mathbf{x}_{n + 1}; \widetilde{\bm \theta}) - \mathbf{\widetilde{K}\Psi}(\mathbf{x}_{n}; \widetilde{\bm \theta})\right\|^{2} + \lambda\left\|\mathbf{\widetilde{K}}\right\|_{\rm F}^{2}}, \label{Loss_Function}
\end{align}
where $\| \cdot \|$ is the Euclidean norm, $\| \cdot \|_{\rm F}$ is the Frobenius norm and $\lambda \in \mathbb{R}$ is a regularization parameter. 
\par
The minimization problem is solved using an iterative method, in which we perform two consecutive procedures at each iteration step. 
First, we fix the parameters $\widetilde{\bm \theta}$ and obtain an optimal finite-dimensional approximation of the Koopman operator $\widetilde{\mathbf{K}}$ to minimize the loss function $J$:  
\begin{align}
  \widetilde{\mathbf{K}} = \left(\mathbf{G}(\widetilde{\bm \theta}) + \lambda \mathbf{I}\right)^{+} \mathbf{A}(\widetilde{\bm \theta}),  \label{K_EDMD-DL}
\end{align}
where $\mathbf{I}$ is the identity matrix. 
Then we fix the approximated Koopman operator and optimize the parameters: 
\begin{align}
  \left(\widetilde{\bm \theta}, \widetilde{\bf v}\right)  \leftarrow \left(\widetilde{\bm \theta} + \delta \Delta {\bm \theta} \left(\mathbf{\nabla}_\theta J(\widetilde{\mathbf{K}}, \widetilde{\bm \theta}), \widetilde{\bf v}, {\bm \beta}\right),   \widetilde{\bf v} + \Delta {\bf v}\left(\mathbf{\nabla}_\theta J(\widetilde{\mathbf{K}}, \widetilde{\bm \theta}), \widetilde{\bf v}, {\bm \beta}\right) \right), \label{theta_EDMD-DL}
\end{align}
where $\widetilde{\bf v}$ denotes, if it exists, internal state of the optimizer, $\delta$ is a learning rate, $\Delta {\bm \theta}, \Delta {\bf v}$ describe update rules of the optimizer and ${\bm \beta}$ is a set of hyperparameters other than the learning rate. 
In this study, we used the Adam algorithm~\cite{Goodfellow} for the optimizer and its hyperparameters ${\bm \beta}$ excluding $\delta$ are set as recommended in~\cite{Kingma}. 

Note that, in order to avoid minimizing the loss function $J$ by a trivial solution $\mathbf{\Psi}(\mathbf{x};\widetilde{\bm \theta}) \equiv 0$, some non-trainable functions, such as constant maps and projection maps ${\bm \pi}_k$ need to be included in $\mathbf{\Psi}$.

\subsection{Dictionaries using neural networks \label{NODE}} 
In this subsection, we introduce MLPs and NODEs, which are used as dictionaries in ths study. 
Note that the MLPs and NODEs can be directly used in order to approximate the law of time-evolution ${\bf F}$. However, through the EDMD-DL, non-trivial dynamical properties, which might be uncertain in such black-box approaches, can be made clear as described in Sec.~\ref{sec. koopman}. 

\begin{figure}[htb]
\begin{center}
\hspace*{0.5cm}
\includegraphics[bb=0 0 1000 600,width=8.0cm,clip]{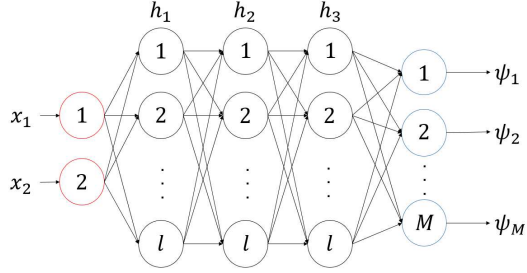}
\caption{MLP-based dictionary for EDMD-DL}\label{MLP}
\end{center}
\end{figure}

Fig.~\ref{MLP} shows an example of the MLP-based dictionary used in the conventional EDMD-DL~\cite{Li}. 
The MLP has three types of layers (input, hidden and output). The state of the $t$-th hidden layer $\mathbf{h}_{t} \in \mathbb{R}^l$ propagates as follows: 
\begin{align}
  \mathbf{h}_{t+1} = \mathbf{h}_{t} + {\bf f}_{\rm MLP}(\mathbf{h}_{t}, {\bm \theta}_{t}). \label{eq. mlpupd}
\end{align}
Here, ${\bf f}_{\rm MLP}$ is some nonlinear transformation and ${\bm \theta}_{t} \subset {\bm \theta}$ is a set of parameters (weights and biases) of the $t$-th layer. 
In this study, we used compositions of an affine transformation mapping and a hyperbolic tangent activation as ${\bf f}_{\rm MLP}$. 
The weights and biases of the affine transformations are initialized using a uniform distribution between $-\sqrt{l}$ and $\sqrt{l}$ in this study. 
The MLP receives vector-valued data $\mathbf{x}_i$s observed from a nonlinear dynamical system and outputs values of the dictionary functions $\psi_{i}$s. 
Then the loss $J$ is obtained and its gradient is calculated using the back-propagation algorithm~\cite{Goodfellow}, that is, $\mathbf{\nabla}_{\bm{\theta}_t} J$ is evaluated sequentially from the output layer to the input layer. In most classes of MLPs, we need to store all the intermediate hidden states ${\bf h}_t$ at the beginning of the back-propagation phase, because ${\bf f}_{\rm MLP}$ might not be invertible. Hence, memory cost of the MLP grows in proportion to its number of layers. 
The dictionary is trained so that it efficiently spans an invariant subspace of the Koopman operator. 
The MLP can be trained in a supervised fashion only with the observed time series. 
%
%

The propagation of the hidden state Eq.~(\ref{eq. mlpupd}) of MLPs can be considered as a discrete-time dynamical system. 
In contrast, NODEs describe the propagation of hidden state $\mathbf{h} \in \mathbb{R}^{l'}$ as a continuous-time dynamical system as follows: 
\begin{align}
  \frac{d\mathbf{h}(t)}{dt} = \mathbf{f}(\mathbf{h}(t), t, {\bm \theta}).  \label{eq. nodeupd} 
\end{align}
The depth of the NODEs is parametrized by a continuous variable, which is an independent variable $t$ of the system of ODEs~(\ref{eq. nodeupd}). 
In this paper, we used the following vector field: 
\begin{align}
  \mathbf{f}(\mathbf{h}(t), t, {\bm \theta}) = ({\bf W}_3\tanh \circ {\bf W}_2 \tanh \circ {\bf W}_1) {(\bf h}(t)), \label{eq. nodevf} 
\end{align}
where ${\bf W}_1 \in \mathbb{R}^{l'\times l''}, {\bf W}_2 \in \mathbb{R}^{l''\times l''}, {\bf W}_3 \in \mathbb{R}^{l''\times l'}$ are matrices independent of $t$ and ${\rm tanh}$ denotes an element-wise operation of a hyperbolic tangent activation. 
Initialization of the components of these matrices is performed using an i.i.d sample with the same normal distribution $\mathcal{N}(0,0.01)$. 
We need not introduce parameters as layer-wise weights and biases, whose numbers increase with the depth of MLPs. 
Thus a parameter-efficient neural network can be realized. 
The NODEs' initial condition is prepared as input at some time instant $t_s$ and its associated output is obtained as a solution of the ODEs evaluated at some $t_e$. The adjoint sensitivity method~\cite{Chen} tells us that $\mathbf{\nabla}_{\bm \theta} J$ can be calculated using 
\begin{align}
  \mathbf{\nabla}_{\bm \theta} J = \int_{t_{1}}^{t_{0}} \mathbf{a}(t)^{\top} \frac{\partial {\bf f}(\mathbf{h}(t), t, {\bm \theta})}{\partial {\bm \theta}} d t, \label{eq. p_sensitivity} 
\end{align}
where $\bf a$ is adjoint variables given by 
\begin{align}
  \frac{d \mathbf{a}(t)}{d t}=-\mathbf{a}(t)^{\top} \frac{\partial f(\mathbf{h}(t), t, {\bm \theta})}{\partial \mathbf{h}}. \label{eq. adj_sensitivity} 
\end{align}
The terminal condition of the adjoint variables is ${\bf a}(t_e) = \mathbf{\nabla}_{{\bm h}(t_e)} J$. 
Once ${\bf h}(t_e)$ and $\mathbf{\nabla}_{{\bm h}(t_e)} J$ is obtained, the equations (\ref{eq. nodeupd},\ref{eq. p_sensitivity},\ref{eq. adj_sensitivity}) can be integrated backward-in-time uniquely if the NODEs (\ref{eq. nodeupd}) admit a unique solution. When compared to the back-propagation in the MLPs, the adjoint sensitivity method does not require the intermediate hidden states to be stored in memory at once, since they can be sequentially evaluated. Thus the NODEs are trained in a memory-efficient manner. 
\begin{figure}[htb]
\begin{center}
\hspace*{0.8cm}
\includegraphics[bb=0 0 1000 600,width=8.0cm,clip]{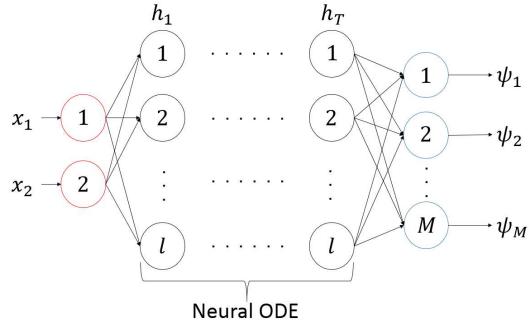}
\caption{NODE-based dictionary for EDMD-DL}\label{NODE_fig}
\end{center}
\end{figure}

\subsection{Proposed method} \label{proposed-method} 
MLPs require many parameters to express dictionaries since the number of them grows in proportion to that of their hidden layers. 
Our goal is to improve parameter efficiency of the EDMD-DL by using NODEs.

In this study, we replace the hidden layers of the MLPs (see Fig.~\ref{MLP}) by the NODEs (see Fig.~\ref{NODE_fig}). The input layer and output layer are the same as in the case of the conventional MLP-based EDMD-DL.
Also, the parameter update scheme shares the same idea as that of the original EDMD-DL~\cite{Li}. 
Algorithm~\ref{alg1} shows how our proposed method is performed. The hyperparameter $\epsilon$ is the tolerated error in the finite-dimensional approximation of the Koopman operator. 

\begin{algorithm}[H]
\caption{EDMD-DL Algorithm using NODEs}              
\label{alg1}                          
\begin{algorithmic}[1]              
\STATE \textbf{function} make\_dictionary
\STATE \ \ \ \ Input $N+1$ data (see Eq.~\eqref{eq. dataset}) to NODEs
\STATE \ \ \ \ Obtain trainable part of $\mathbf{\Psi}(\mathbf{x})$ as output of NODEs
\STATE \ \ \ \ Compute non-trainable part of $\mathbf{\Psi}(\mathbf{x})$
\STATE \ \ \ \ Concatenate the trainable part and nontrainable part \\ \ \ \ \ then define it as \textit{dictionary}
\STATE \ \ \ \ \textbf{return} \textit{dictionary}
\STATE \textbf{end function}
\item[]

\WHILE{$J(\widetilde{\mathbf{K}}, \widetilde{\bm \theta}) > \epsilon$}
\STATE make\_dictionary\
\STATE Compute $\mathbf{G}$, $\mathbf{A}$, and $\widetilde{\mathbf{K}}$ (See Eqs.~(\ref{G}, \ref{A}, \ref{K_EDMD-DL}))
\STATE Optimize $J(\widetilde{\mathbf{K}}, \widetilde{\bm \theta})$ with respect to $\bm \theta$ using the Adam
\ENDWHILE
\item[]

\STATE Define $\mathbf{K}$ as $\mathbf{K} = \widetilde{\mathbf{K}}$
\STATE Obtain $\mu_k$s, ${\bm \xi}_k$s, and ${\bm \zeta}_k$s from $\mathbf{K}$
\STATE Find $\mathbf{B}$ (See Eq.~\eqref{B})
\STATE Compute $\mathbf{m}_{k}$s
\STATE Compute $\phi_{k}$s
\item[]
\end{algorithmic}
\end{algorithm}

\section{Results} 
In this section, we evaluate parameter efficiency of the NODE-based EDMD-DL (proposed method) and compare it to that of the MLP-based EDMD-DL (conventional method). 
Time series prediction tasks using Eq.~(\ref{predictive_Koopman}) are considered for two nonlinear dynamical systems, the Duffing equation and the Kuramoto-Sivashinsky equation.

\subsection{Performance metrics and evaluation} \label{error} 
In this subsection, we introduce two metrics about the error in the EDMD-DL. One is the trajectory reconstruction error, and the other is the error in eigenfunction approximation.
\subsubsection{Trajectory reconstruction error} \label{E_traj} 
In this section, we perform predictions on nonlinear time series. We define the error between the predicted trajectory and the actual trajectory as:
\begin{align}
E_{\rm recon} =  \sqrt{\frac{1}{N} \sum_{n = 1}^{N}{\| \mathbf{x}(n) - \tilde{\mathbf{x}}(n) \|^2}}, \label{recon_E_eq}
\end{align}
where $\tilde{\mathbf{x}}(n)$ denotes the trajectory predicted using Eq.~(\ref{predictive_EDMD}). 
Note that errors are computed using initial conditions that are not seen in the training phase. 
\subsubsection{Error in eigenfunction approximation} 
The approximation error for the $j$-th eigenfunction ($j = 1, 2, ..., M$) is defined as:
\begin{align}
E_{j} =  \|\widetilde{\phi}_{j} \circ \mathbf{F} - \widetilde{\mu}_{j} \widetilde{\phi}_{j}\|_{L^2(\mathcal{M},\:\mu)}, \label{eigen_E_eq}
\end{align}
where $\widetilde{\mu}_{j}$ and $\widetilde{\phi}_{j}$ are the eigenvalue and the eigenfunction obtained by Algorithm~\ref{alg1}, respectively. 
In Eq.~\eqref{eigen_E_eq}, $\|\cdot\|_{L^2(\mathcal{M},\: \mu)}$ represents $L^2(\mathcal{M}, \mu)$ norm. 
In this study, $\mu$ is a uniform distribution on a subset of $\mathcal{M}$ under consideration. 
Each eigenfunction $\widetilde{\phi}_{j}$ is normalized by its $L^2(\mathcal{M}, \mu)$ norm. 
In this numerical experiment, we approximately evaluate $L^2(\mathcal{M}, \mu)$ norm by using a Monte-Carlo integration:
\begin{align}
  \widetilde{\phi}_{j} \leftarrow \frac{1}{\sqrt{a}}\widetilde{\phi}_{j}, \label{normalization}
\end{align}
where $a=\frac{1}{I}\sum_{i = 1}^{I}|\widetilde{\phi}_{j}(\bar{\mathbf x}(i))|^{2}$, $I$ is a sample size, and $\bar{\mathbf x}(i)$ is sampled from the uniform distribution. 
In this study, we set $I = 10000$.
\par
In Eq.~\eqref{eigen_E_eq}, we approximately evaluate the error by using a Monte-Carlo integration:
\begin{align}
E_{j} \approx \sqrt{\frac{1}{I} \sum_{i = 1}^{I} \left|\widetilde{\phi}_{j} \circ \mathbf{F}(\bar{\mathbf x}(i)) - \widetilde{\mu}_{j} \widetilde{\phi}_{j}(\bar{\mathbf x}(i))\right|^{2}}. \label{eigen_E}
\end{align}
\par
We use an averaged error as a metric for the eigenfunction error:
\begin{align}
E_{\rm eigen} = \frac{1}{M} \sum_{j = 1}^{M} E_{j}. \label{eigen_E_avg}
\end{align}
This quantifies the accuracy of the approximation of the Koopman operator. 
\subsubsection{Performance evaluation} \label{sec. eval}
We first fix the number of trainable parameters of the MLP. Secondly, the EDMD-DL is performed using the MLP and the two performance metrics $E_{\rm recon}, E_{\rm eigen}$ are evaluated. Then the NODE-based EDMD-DL is performed with various values of $l''$ introduced in Sec.~\ref{NODE}. $l''$ controls the number of the trainable parameters of the NODEs. For each performance metric, we compare the number of the trainable parameters of the DNNs when the metric is aligned. 
The less the number of the trainable parameters which produce the same metric, the more parameter-efficient the method is. 

\subsection{Nonlinear time series prediction and measurement of parameter efficiency} 
%

\subsubsection{Duffing equation} \label{Duffing} 
The Duffing equation is described by the following equation:
\begin{align}
\ddot{x}_1 = - \gamma \dot{x}_1 - x_1(\beta + \alpha x_1^2). \label{Duffing_eq}
\end{align}
We set its parameters as $\alpha = 1$, $\beta = -1$, and $\gamma = 0.5$, at which the system shows multi-stability: $x_1$ converges to $1$ or $-1$ depending on the initial condition. 
The system is time-discretized by a step size $\Delta t = 0.25$. 
\par
The predictions are made as follows.
First, we sample 1000 random initial conditions from the uniform distribution on $[-2,2]^2$. 
Each initial condition is time-integrated over 10 steps then the data set Eq.~(\ref{eq. dataset}) of $N=10000$ is prepared.
Second, we train the DNNs using the data set and obtain the Koopman eigenpairs and modes using the Algorithm~\ref{alg1}. 
The matrix $\mathbf{B}$ is set so that $\mathbf{g}$ is the identity map.  
Then the predictions are made using Eq.~(\ref{predictive_EDMD}) on newly sampled initial conditions. 
In the training phase, we prepare $(M - 3)$ trainable dictionary outputs and 3 non-trainable ones, which are one constant map and two coordinate projection maps.
\par
\par
The hyperparameters are set as follows: the number of dictionary elements $M=25$, the width of the MLP (resp. NODEs) $l=170$ (resp. $l'=120$), the depth of the hidden layer in the MLP is $3$, the Tikhonov regularization parameter $\lambda = 0.01$, the tolerance $\epsilon = 30$ and the learning rate $\delta = 0.001$.
The Duffing equation (\ref{Duffing_eq}) and the NODEs (\ref{eq. nodeupd}) are solved using the Dormand-Prince 5(4) method~\cite{dopri} with an allowable absolute (resp. relative) error $10^{-9}$ (resp. $10^{-7}$). 
\par
\begin{figure}[H]
 \begin{minipage}{0.5\hsize}
  \begin{center}
\includegraphics[bb=0 0 1100 350,width=16cm,clip]{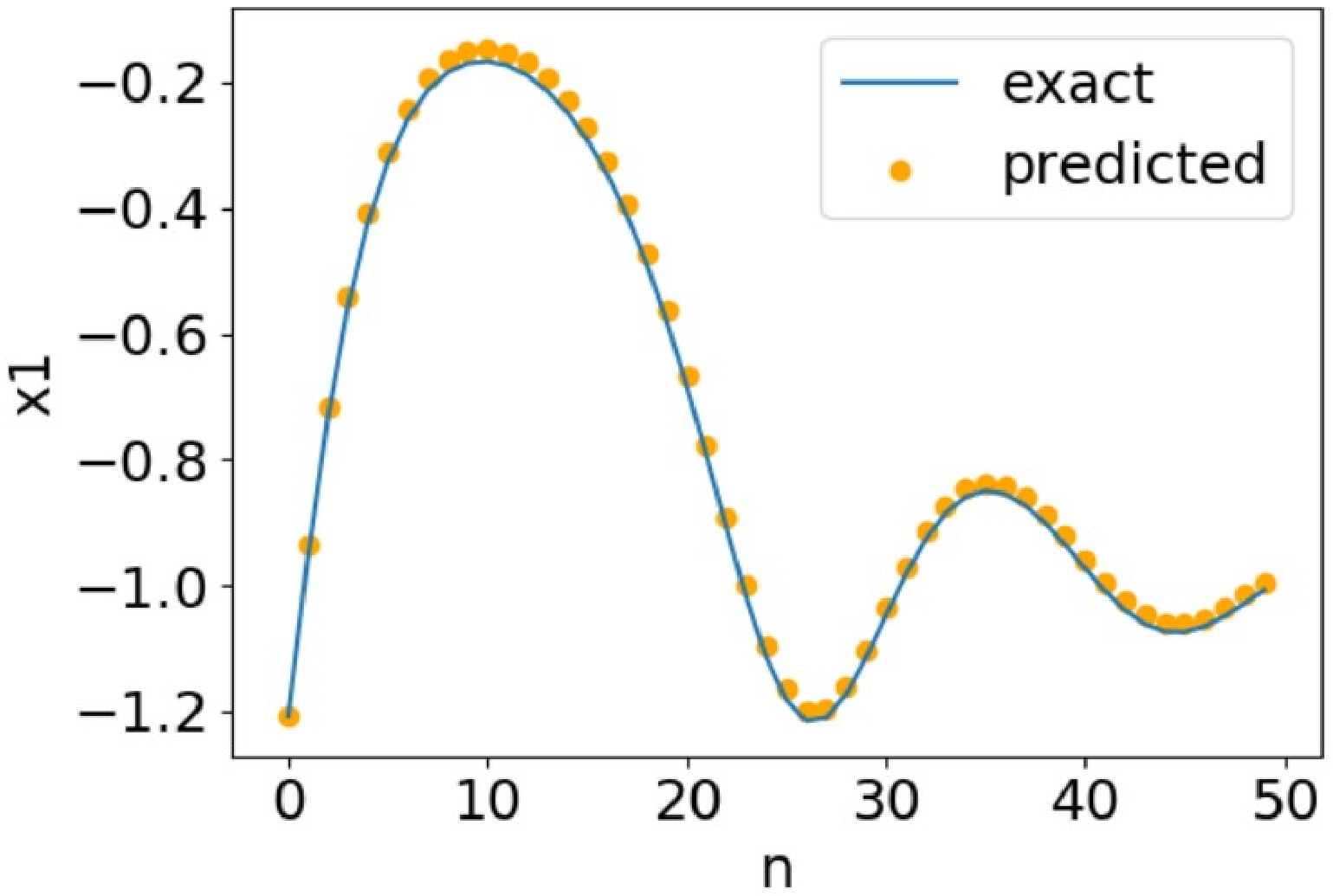}
  \end{center}
  \label{figure_4_3}
 \end{minipage}
 \begin{minipage}{0.5\hsize}
  \begin{center}
\includegraphics[bb=0 0 1100 350,width=16cm,clip]{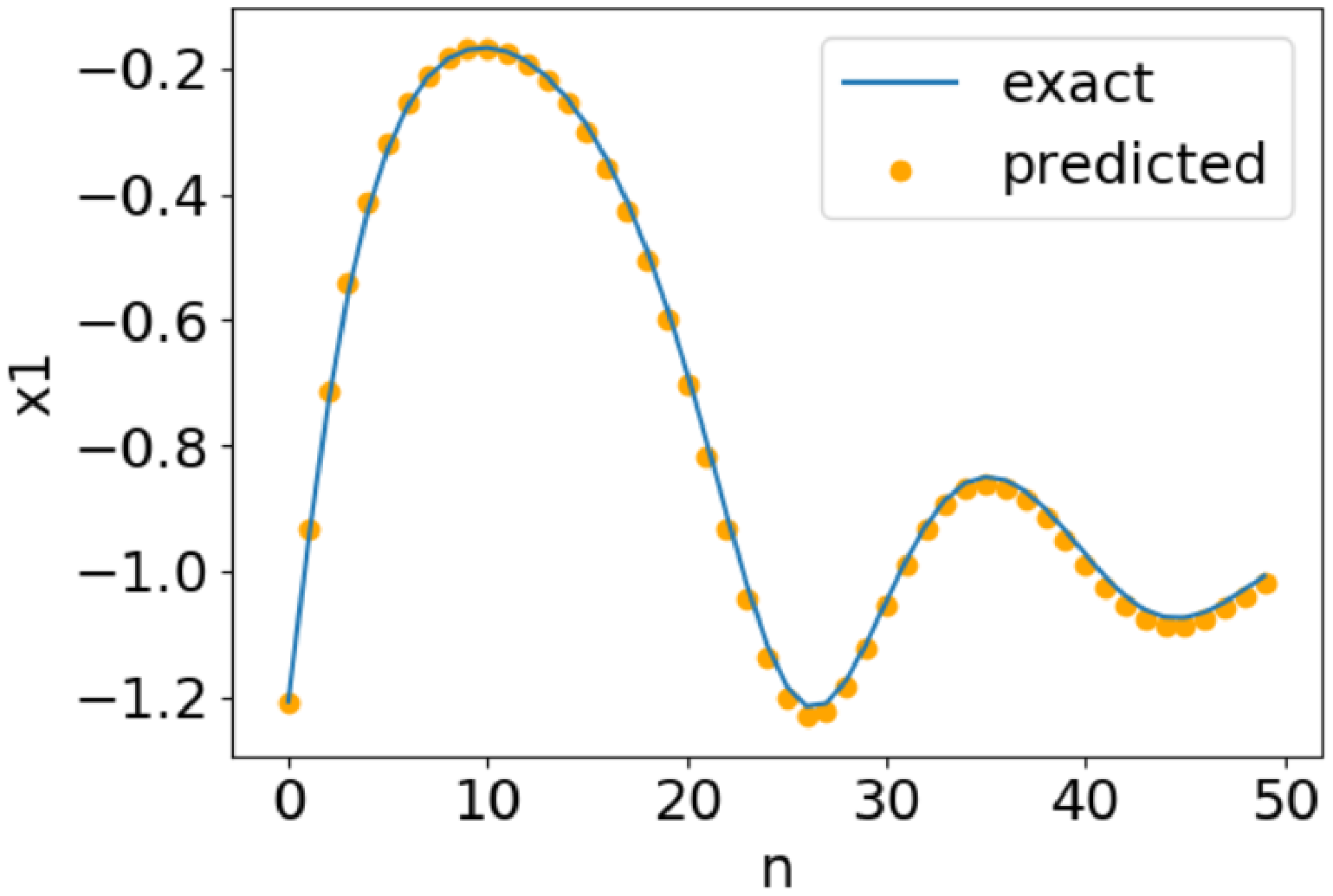}
  \end{center}
  \label{figure_4_4}
 \end{minipage}
\caption{
Prediction of time series produced by the Duffing equation~\eqref{Duffing_eq} using the EDMD-DL. The dotted orange lines indicate predictions and the solid blue lines are exact data. Upper left (resp. upper right) panel: prediction by the conventional (resp. proposed) method when $x_1$ converges to $1$. Lower left (resp. lower right) panel: prediction by the conventional (resp. proposed) method when $x_1$ converges to $-1$. In all figures,  
both the conventional method and the proposed one give good predictions.} \label{traj_Duffing}
\end{figure}

Fig.~\ref{traj_Duffing} shows the prediction results obtained by the MLP-based EDMD-DL and the NODE-based one, respectively. 
In Fig.~\ref{traj_Duffing}, the horizontal axis is the number of steps $n$, the vertical axis is the coordinate $x_1$, the solid blue lines indicate actual data and the dotted orange lines show EDMD-DL prediction. 
The NODE-based EDMD-DL shows good prediction performance as that of the MLP-based one.
\begin{figure}[H]
 \begin{minipage}{0.5\hsize}
  \begin{center}
\includegraphics[bb=0 0 1100 350,width=16cm,clip]{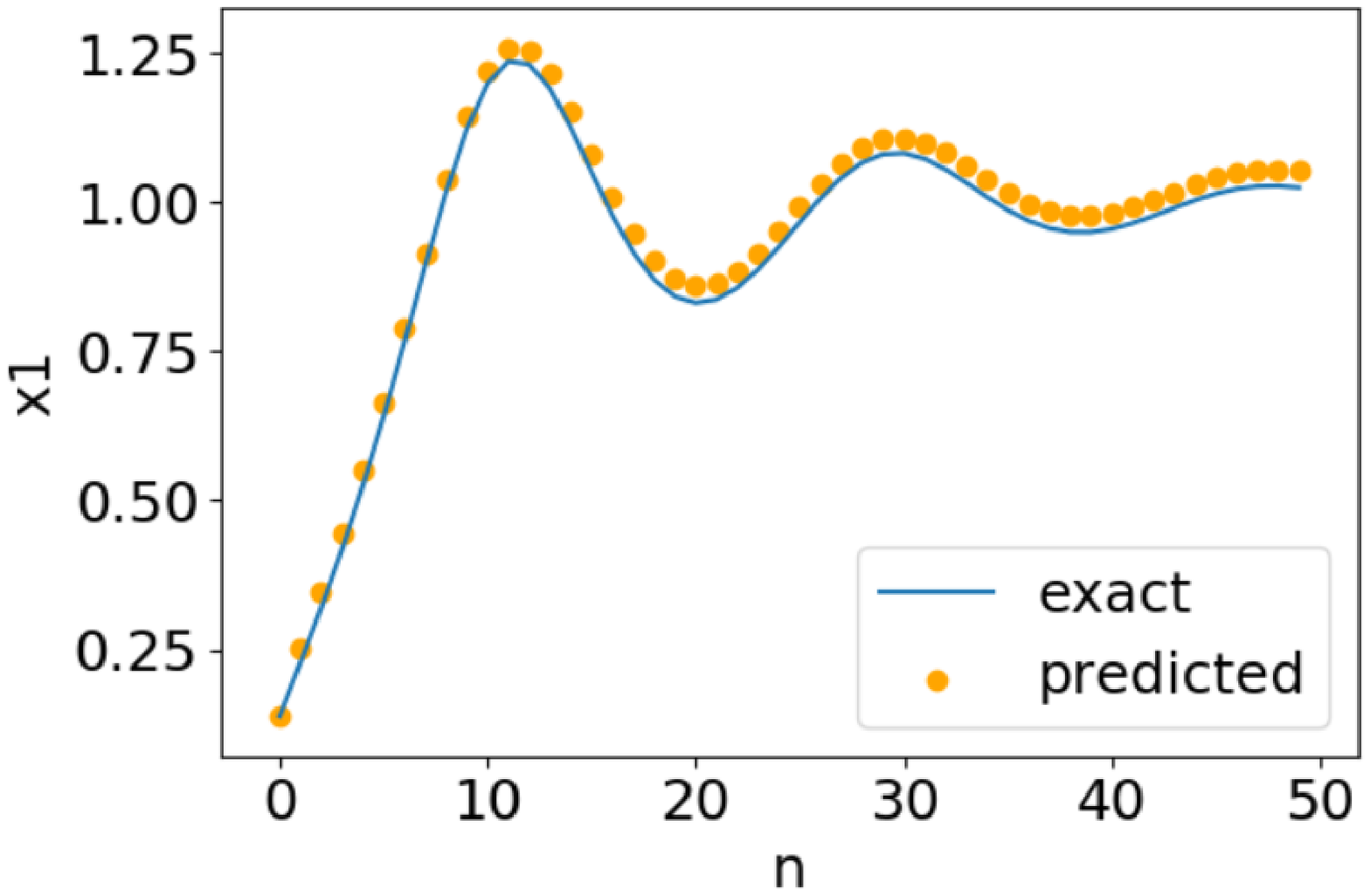}
  \end{center}
  \label{figure_4_1}
 \end{minipage}
 \begin{minipage}{0.5\hsize}
  \begin{center}
\includegraphics[bb=0 0 1100 350,width=16cm,clip]{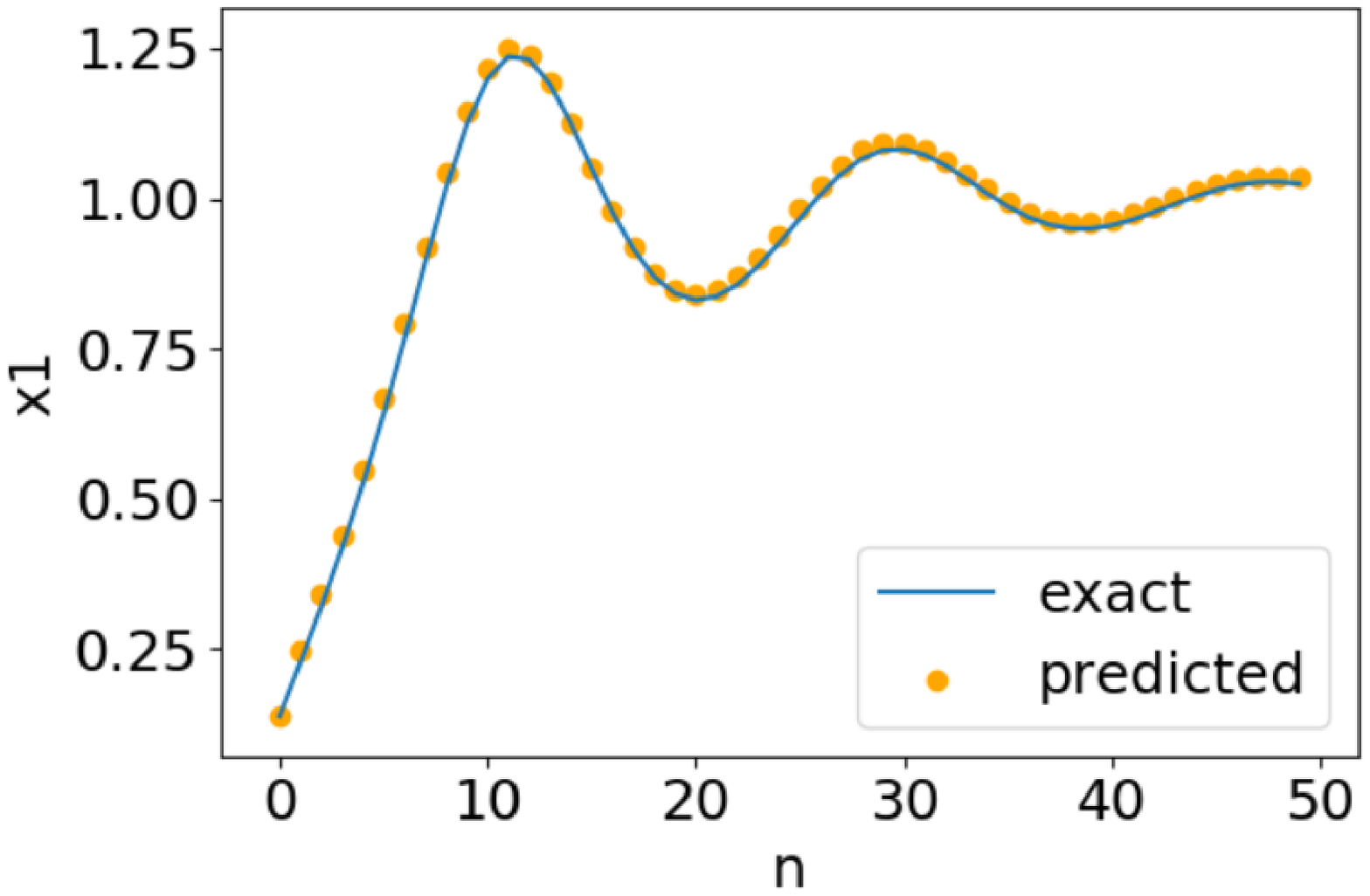}
  \end{center}
  \label{figure_4_2}
 \end{minipage}
\end{figure}
Table~\ref{Duffing_recon_table} shows the trajectory reconstruction errors and their associated number of parameters. 
Table~\ref{Duffing_eigen_table} compares the parameter efficiency in terms of the eigenfunction error. 
It is seen that, in both cases, the proposed method is more than two and a half times as parameter-efficient as the conventional one.
\begin{table}[htb]
  \begin{center}
    \caption{Trajectory reconstruction error in predictions of Eq.~\eqref{Duffing_eq} and their associated number of parameters}
  \begin{tabular}{|c|c|c|c|} \hline
     & $E_{\rm recon}$ (Convergence to $x_1 = 1$) & $E_{\rm recon}$ (Convergence to $x_1 = -1$) & number of parameters \\ \hline
    conventional & $7.8 \times 10^{-3}$ & $5.4 \times 10^{-3}$ & 62,412 \\ \hline
    proposed & $6.9 \times 10^{-3}$ & $4.5 \times 10^{-3}$ & 23,895 \\ \hline
  \end{tabular}
  \label{Duffing_recon_table}
  \end{center}
\end{table}
\par
\begin{table}[htb]
  \begin{center}
    \caption{Eigenfunction error in predictions of Eq.~\eqref{Duffing_eq} and their associated number of parameters}
  \begin{tabular}{|c|c|c|} \hline
     & $E_{\rm eigen}$ & number of parameters \\ \hline
    conventional & 0.90 & 62,412 \\ \hline
    proposed & 0.87 & 22,605 \\ \hline
  \end{tabular}
  \label{Duffing_eigen_table}
  \end{center}
\end{table}

\par
\begin{figure}[H]
\begin{center}
\includegraphics[width=\textwidth, clip]{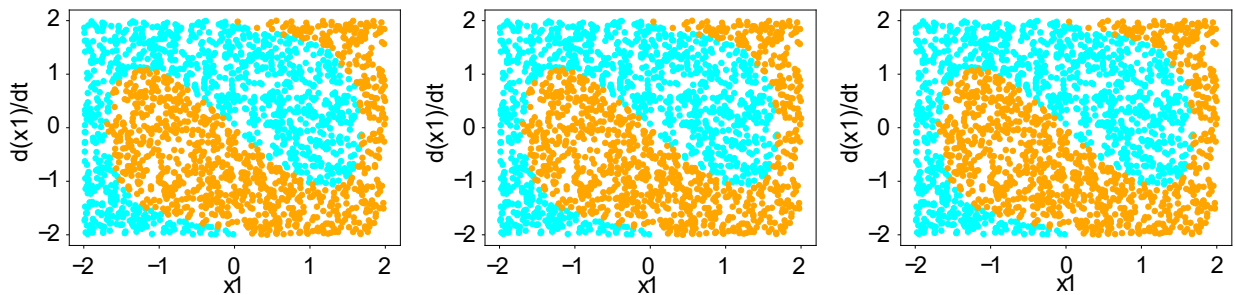}
\end{center}
\caption{ Classification of the limiting equilibrium state of the Duffing equation~\eqref{Duffing_eq} using the EDMD-DL. The orange dots denote initial conditions that converge to $x_1 = -1$ and the blue dots represent those to $x_1 = 1$. Left panel: results obtained by the conventional method. Middle panel: results obtained by the proposed method. Right panel: exact data} \label{converge_MLP_predicted}
\end{figure}

The Koopman eigenfunctions provide not only trajectory-wise predictions for nonlinear dynamical systems but also information on their global phase space structure. For example, it can be seen that a Koopman eigenfunction corresponding to a unit eigenvalue serves as an indicator of basins of attraction. Thus, a good approximation of the Koopman operator solves a classification task of initial conditions in terms of their limiting behavior. Fig.~\ref{converge_MLP_predicted} shows the classification results for the Duffing equation using 2000 newly sampled initial conditions as test inputs. Test outputs are defined by whether $x_1(49\Delta t)$ is closer to $1$ or $-1$. Similarly, the classification outcome is determined using $\widetilde{x}_1(49\Delta t)$. It can be seen that our proposed method successfully solves this classification task. 
%
\noindent
\par
Table~\ref{Duffing_table_separation} shows the minimum number of parameters that the above-mentioned 2000 samples are classified as in Fig.~\ref{converge_MLP_predicted}. 
The parameter efficiency of our proposed method is more than twofold that of the conventional one. 
\begin{table}[H] 
  \begin{center}
    \caption{Minimum number of parameters that shows the classification outcome of the limiting equilibria of the Duffing equation as in Fig.~\ref{converge_MLP_predicted}}
  \begin{tabular}{|c|c|} \hline
      & number of parameters \\ \hline
    conventional & 14,296 \\ \hline
    proposed & 6,054 \\ \hline
  \end{tabular}
  \label{Duffing_table_separation}
  \end{center}
\end{table}
\noindent

\subsubsection{Kuramoto-Sivashinsky equation} \label{Kuramoto-Sivashinsky} 
The Kuramoto-Sivashinsky equation, which exhibits spatio-temporal chaos~\cite{Kuramoto}, is described as: 
\begin{align}
  \frac{\partial u}{\partial t} = - \frac{\partial^2 u}{\partial x^2} - \frac{\partial^4 u}{\partial x^4} - u\frac{\partial u}{\partial x},
\label{K-S_eq}
\end{align}
where $x \in \mathbb{R}$ and $t \in \mathbb{R}$. We assume that the periodic boundary condition $u (x, t) = u (x + L, t)$, where $L=16$, is satisfied. 
The space is divided into $N_x = 128$ evenly spaced grids, i.e., the space step size is $L/N_x = 0.125$ and the discrete time step size is set $\Delta t = 0.005$.  We sample 100 random initial conditions from the uniform distribution on $[-4,4]^{N_x}$ and each initial condition is time-integrated over 100 steps, that is, the data set Eq.~(\ref{eq. dataset}) of $N=10000$ is prepared. The identity map is used as the observable. We prepare a dictionary composed of $(M - 129)$ trainable elements and 129 non-trainable ones. A constant map and standard projections constitute the set of the non-trainable elements. 
\par
The hyperparameters of the training are set as follows: the number of dictionary elements $M=151$, the width of the MLP (resp. NODEs) $l=303$ (resp. $l'=156$), the depth of the hidden layer in the MLP is $3$, the Tikhonov regularization parameter $\lambda = 0.01$, the tolerance $\epsilon = 30$ and the learning rate $\delta = 0.001$. 
%
When we solve the Kuramoto-Sivashinsky equation, the state space is discretized using the central difference scheme and time integration is performed using the explicit Euler method. The NODEs (\ref{eq. nodeupd}) are solved using the Dormand-Prince 5(4) method of the same tolerable errors as those of the Duffing case. 
\par
\begin{figure}[H]
\begin{center}
\includegraphics[width=\textwidth,clip]{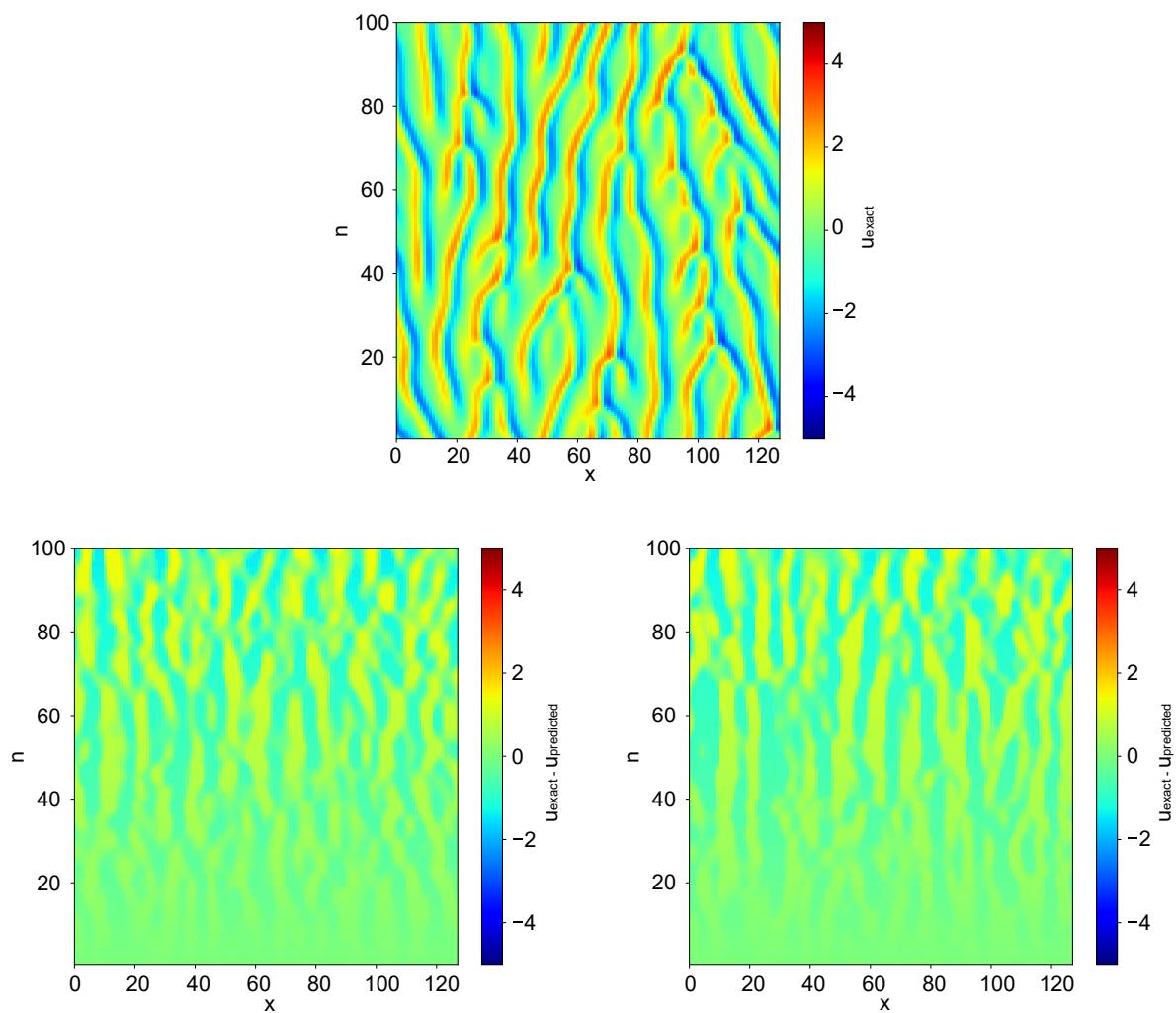}
\end{center}
\caption{ Top panel: exact time series of the Kuramoto-Sivashinsky system \eqref{K-S_eq}. Bottom panels: Errors between the exact time series and those that reconstruct it using the EDMD-DL methods. The lower left (resp.~right) panel shows the result for the conventional (resp.~proposed) method.} 
\label{traj_K-S}
\end{figure}

Illustrations of the predictions of the Kuramoto-Sivashinsky system is shown in Fig.~\ref{traj_K-S}. In the top panel, the exact time series obtained by direct numerical simulation of \eqref{K-S_eq} is plotted and the bottom panels show the errors between the exact time series and those that reconstruct it using the EDMD-DL.  
The horizontal axis indicates the spatial coordinate $x$ and the vertical axis denotes the number of time steps $n$. The prediction made by the NODE-based EDMD-DL shows good accuracy as that of the MLP-based one. 

Tables~\ref{K-S_recon_table},\ref{K-S_eigen_table} show the parameter efficiency in terms of the reconstruction error and the eigenfunction error, respectively. In both metrics, the proposed method is more than three times as parameter-efficient as the MLP-based one.
\begin{table}[htb]
  \begin{center}
    \caption{Trajectory reconstruction error in predictions of Eq.~\eqref{K-S_eq} and their associated number of parameters}
  \begin{tabular}{|c|c|c|c|} \hline
     & $E_{\rm recon}$ & number of parameters \\ \hline
     conventional & $1.13$ & 191,821 \\ \hline
     proposed & $1.09$ & 57,504 \\ \hline
  \end{tabular}
  \label{K-S_recon_table}
  \end{center}
\end{table}
\par
\begin{table}[htb]
  \begin{center}
    \caption{Eigenfunction error in predictions of  Eq.~\eqref{K-S_eq} and their associated number of parameters}
  \begin{tabular}{|c|c|c|} \hline
     & $E_{\rm eigen}$ & number of parameters \\ \hline
    conventional & 0.85 & 191,821 \\ \hline
    proposed & 0.84 & 55,166 \\ \hline
  \end{tabular}
  \label{K-S_eigen_table}
  \end{center}
\end{table}

\section{Conclusion} 
Time series analysis of nonlinear phenomena is an important but challenging problem. 
We can investigate the nonlinear dynamic behavior by linear techniques through an approximation of the Koopman operator, which can be obtained in data-driven fashions using the extended dynamic mode decomposition and its variants. 
It is desired to develop computationally-efficient method to perform the EDMD-type algorithms.  
In this study, we developed a parameter-efficient method to provide the approximation of the Koopman operator by incorporating the neural ordinary differential equations as a dictionary for the EDMD-DL. 
The proposed method was illustrated using the Duffing and the Kuramoto-Sivashinsky systems. Their characteristic complexities such as multi-stability and spatio-temporal chaos were quantitatively addressed with good accuracy by the proposed method. The proposed method yielded improved parameter efficiency over the conventional EDMD-DL. 
The proposed method would be applied to complex nonlinear dynamical phenomena in the real-world such as atmospheric changes, spatio-temporal regulations of biological networks, and stock price fluctuations. Moreover, the interpretability of the proposed method, derived from the Koopman operator framework, would facilitate understanding of such intriguing phenomena. 

\section*{Acknowledgments}
This study is supported by JST CREST Grant Number JPMJCR18K2 and JST Moonshot R\&D Grant Number JPMJMS2021, Japan.

\end{document}